\documentclass[10pt, conference, compsocconf]{IEEEtran}
\newcommand{\eg}{{\it e.g., }}
\newcommand{\etal}{{\it et~al. }}
\newcommand{\ie}{{\it i.e., }}

\usepackage{times}
\usepackage{url}
\usepackage{latexsym}
\usepackage{color}
\newcommand{\name}{\emph{SensPick}}
\usepackage{esvect}
\usepackage{amssymb,float,varioref,balance}
\usepackage{graphicx}
\usepackage{epsfig}
\usepackage{tabularx,multirow,booktabs,blindtext}
\usepackage{balance}

\title{SensPick: \underline{Sens}e \underline{Pick}ing for Word Sense Disambiguation}

\author{\IEEEauthorblockN{Sm Zobaed\IEEEauthorrefmark{1},
Md Enamul Haque\IEEEauthorrefmark{2}, 
Md Fazle Rabby\IEEEauthorrefmark{1}, and
Mohsen Amini Salehi\IEEEauthorrefmark{1}}

\IEEEauthorblockA{\IEEEauthorrefmark{1}School of Computing \& Informatics \\
University of Louisiana at Lafayette, Louisiana 70504, USA \\
}
\IEEEauthorblockA{\IEEEauthorrefmark{2}Spencer Center for Vision Research\\
Stanford University School of Medicine, Palo Alto, CA 94303, USA\\Email: \{sm.zobaed1, enamul, mdfazle.rabby1, amini\}@louisiana.edu
}
}

\renewcommand\vec[1]{\overrightarrow{#1}}
\newcommand\cev[1]{\overleftarrow{#1}}

\begin{document}
\maketitle
\begin{abstract}
Word sense disambiguation (WSD) methods identify the most suitable meaning of a word with respect to the usage of that word in a specific context. Neural network-based WSD approaches rely on a sense-annotated corpus since they do not utilize lexical resources. 
In this study, we utilize both context and related gloss information of a target word to model the semantic relationship between the word and the set of glosses. 
We propose \name, a type of stacked bidirectional Long Short Term Memory (LSTM) network to perform the WSD task. 
The experimental evaluation demonstrates that \name~outperforms traditional and state-of-the-art models on most of the benchmark datasets with a relative improvement of $3.5\%$ in F-1 score. While the improvement is not significant, incorporating semantic relationships brings \name~in the leading position compared to others. 

\end{abstract}

\section{Introduction}
\label{sec:introduction}
Natural languages consist of complex structures and connotations based on various contextual usage. 
One complex challenge in studies related to automated language understanding is to identify appropriate contextual expression of a word having multiple meanings or senses.
For instance, the word ``play'' implies ``activity'' and ``drama'' in ``Sir Donald George Bradman used to \emph{play} cricket for Australia'' and ``Hamlet is one of the famous tragic \emph{plays} of all time'' sentences, respectively.
Although humans effortlessly perceive the context-specific meaning of an ambiguous word utilizing prior knowledge, the disambiguation of an ambiguous word is a challenging task for a machine.

Automated language processing algorithms require the processing of unstructured plain-text data and transformation into certain data structures that provide flexibility to determine the inherent meaning. 
The ability to computationally determine the exact sense of an ambiguous word from a sentence or phrase is known as Word Sense Disambiguation (WSD)~\cite{navigli2009word,pal2019novel}.  
WSD methods are utilized in a wide variety of real-world applications such as machine translation, word processing, and information retrieval (IR) domain~\cite{kaageback2016word}. 
In addition, WSD performs a substantial role in the query processing (\ie query expansion) of a search engine to facilitate the retrieval of relevant search results. 
Further, the relevancy of a search result depends on the appropriate query expansion. 
Without utilizing WSD in query expansion, search results could be impacted by non-relevant information. 
However, in the natural language processing (NLP) domain, WSD is a long-standing open challenge problem~\cite{chaplot2018knowledge}. 
Several studies are performed on WSD tasks with respect to knowledge-based, supervised, and unsupervised models. 
The knowledge-based WSD models leverage external knowledge bases such as thesaurus and semantic dictionaries to obtain sense definitions of the words of a given sentence~\cite{pal2019novel,lesk1986automatic,tang2010semi}.

Unsupervised models create sense-tagged clusters of training sentences by utilizing different clustering algorithms such as \emph{k}-means, hierarchical clustering, or their modified versions~\cite{karim2020efficient}. 
To identify an appropriate cluster for a test sentence, such models calculate the closeness score of the sentence with respect to the clusters. 
The sense of the cluster which yields the highest closeness score is selected as the appropriate sense for the test sentence. 
Note that a few major challenges such as performing proper clustering, number of cluster approximation, and tagging senses across the clusters are used as constraints in these models. 
Supervised models generally train multiple classifiers with a manually tailored feature set. Although these models yield state-of-the-art performance compared to others, they are not sufficiently scalable since they are costly in terms of time overhead~\cite{chen2013applying}. 

Generally, neural models consider the local context of the target word without leveraging lexical resources such as WordNet and BabelNet~\cite{miller1995wordnet, navigli2012babelnet}. 
Such lexical resources offer gloss information that denotes all possible definitions of a word considering target sense. 
Recent studies demonstrate that the improvement in the WSD task is achievable by incorporating possible gloss information of the target word~\cite{basile2014enhanced,chaplot2018knowledge}. 
Hence, we leverage the gloss information of the target word into a neural network-based model to achieve improvement in WSD task accuracy.
Since the enriched gloss set is the key to achieve better accuracy, we propose to expand the gloss set by considering the semantics of the target word. 
To this end, we consider the expansion of the gloss set by incorporating the sense-level definition of hypernyms and hyponyms (depends on availability) of the target word.

In this study, we propose a novel neural network-based model \emph{\name} to perform the WSD task with improved accuracy. 
\emph{\name}~combines the target word and its gloss set to build an adequate semantic relationship between the glosses and the target word. 
To build the model, we propose to leverage a specific form of long short-term memory network (LSTM)~\cite{gers2000learning} based architecture, namely stacked bi-directional LSTM since LSTM is widely adopted in sequence learning neural network domain (\eg Seq2seq models).    

In summary, the main contributions of this research are enlisted as follows:
\begin{itemize}
    \item Unlike knowledge-based models, the proposed approach exploits both sense-tagged data and lexical resources. 
    \item To the best of our knowledge, there is no study that solves word sense disambiguation using
    stacked bi-directional LSTM that generates context and gloss vectors to build a WSD task performer model.
    \item Based on the proper evaluation on \texttt{All-Words} WSD benchmark datasets, \emph{\name}~outperforms the other state-of-the-art works.
\end{itemize}

\section{Related Works}
In this section, we discuss prior WSD research that have influenced the proposed approach.
\subsection{Knowledge-based Models}
Knowledge-based models rely on the gloss information of a target word and lexical resources to disambiguate the target word. They do not require any sense annotated dataset that is utilized in various supervised or neural network-based models.

These models utilize lexical resources such as WordNet, BabelNet to leverage gloss information for the WSD task. A lexical resource is a semantic graph whose vertices and edges of the graph are synsets (set of words that denote the same senses) and semantic relation metrics respectively. 
One of the popular models of the knowledge-based is the Lesk algorithm~\cite{lesk1986automatic} that calculates the overlap or distributional similarity between the context of the target word and its gloss set. Likewise, most of the variants of the knowledge-based models follow the same principle. Additionally, the distributional similarity has also been varied in different works~\cite{camacho2016large}. Another popular variant of the knowledge-based is semantic graph-based models~\cite{guo2010combining,moro2015semeval} where
a graph representation of the input text is created to
utilize various graph algorithms over the
given representation (\eg PageRank~\cite{brin1998anatomy}) to disambiguate the target word.


\subsection{Supervised Models}
Supervised models train various distinguishable features that are extracted from the manually sense-annotated dataset. These features are generally extracted considering the information provided by the co-occurring words of
the target word~\cite{navigli2009word},~\cite{kaageback2016word},~\cite{correa2018word}. Recently, instead of using such types of features, different word embedding techniques (\eg Word2Vec~\cite{mikolov2013efficient}, GloVe~\cite{pennington2014glove}, BERT~\cite{devlin2018bert}) are utilized to obtain more complex features~\cite{luo2018leveraging}. These features are considered as input to train a linear classifier. However, there are dedicated classifiers for each unique word of a given sentence and we need to train \textit{n} number of different classifiers for \textit{n} words in a sentence that cause a plausible scalability issue. Specifically, for \texttt{All-words} WSD task that asks to disambiguate every polysemous word in the sentences~\cite{raganato2017neural} would require a number of different classifiers~\cite{luo2018incorporating}.
\subsection{Neural network-based Models}
To overcome the scalability issue of the supervised models, various neural network-based unified models have been proposed. 
Such models leverage bi-directional long short-term memory networks to build a unified classifier that shares trainable model parameters among the polysemous words~\cite{raganato2017neural}. 
Raganato \etal convert the WSD task into a sequence labeling
task~\cite{raganato2017neural}. 
In \cite{yang2019leveraging}, Yang \etal explain
that incorporating knowledge and labeled data into
a unified neural model can achieve better performance
than others that only learn from the large amount of sense-annotated data.

In recent research works, we have studied that various neural network-based memory networks (\ie LSTM) obtain the state-of-the-art results in a wider variant of NLP tasks
such as sentiment analysis (\cite{wang2016dimensional,baziotis2017datastories}), text summarization (\cite{song2019abstractive,liu2018generative}). 

LSTM is used to learn
more distinguishable representation of lengthy sentential contexts
~\cite{melamud2016context2vec},~\cite{yuan2016semi}. At first, an LSTM model
is trained to learn the context vector for each sense in the training dataset. 
If the context vector of a sense is closer to the target word’s context vector, the model selects the sense as the desired sense.

Motivated by the successful implementation of LSTM in the WSD task, we propose to leverage multi-layered (aka stacked) bi-directional LSTM with the required adaptation so that the model can appropriately detect the underlying closeness between the context and gloss set of the target word.



\section{Background}
In this section, we explain the underlying techniques that are associated to \name.
\subsection{Bi-directional LSTM}
LSTM is a gated type variant of recurrent neural network (RNN) that is introduced~\cite{rumelhart1986learning} to efficiently capture long term dependencies in sequence modeling. 
The underlying technique allows the model to copy the state between timesteps
and does not constrain the state through non-linearity~\cite{kaageback2016word}. Unlike the traditional RNN model which works based on logistic functions, the LSTM model utilizes multiplicative gates that compute the gradients better. 
One of the variants of LSTM named bi-directional LSTM (BiLSTM), where the state at each time step incorporates the state of two LSTMs enunciated as forward and backward LSTM. Forward and backward LSTM goes from left to right direction and vice versa.
In the case of WSD, the state has to utilize the information about the preceding and succeeding words of the target word in a sentence. Based on our investigation, prior research motivate us to utilize BiLSTM in the preceding and succeeding parts to correctly capture the context.

\subsection{Stacked LSTM}
Stacking is an extension of LSTM that is comprised of multiple layers where each layer accommodates multiple memory cells.
Stacking results in the LSTM model deeper which means that the addition of layers increases the abstraction level of input observations over time~\cite{pascanu2014construct}.
Stacked LSTMs or deep LSTMs are introduced by Graves \etal in their application of LSTMs to speech recognition, beating a benchmark on a challenging standard problem~\cite{pascanu2014construct}. 
The authors prioritize the implementation of the depth of a network over the number of memory cells in a given layer.

\subsection{Word Embedding}
Word embedding is a commonly used method to represent words as real-valued vectors in a semantically meaningful predefined space.
Commonly, word embedding is trained on large amounts of data in an unsupervised manner that yields vectors by extracting
distinguishable syntactic and semantic features about words. The resultant vectors are dense since the number of features are much smaller than the total number of considered words.
 Without leveraging word embedding, we would require millions of dimensions for sparse word representations (\ie one-hot encoding) that leads to poor training. Hence, word embedding is used to
initialize the input layer of a neural network to build the NLP model.
Global vectors for word representation (GloVe) is a widely adopted word embedding technique that combines a log-linear model and
co-occurrence statistics to efficiently capture global statistics~\cite{kaageback2016word}.

\section{Sense Picking for WSD}
\begin{figure*} [h]
    \centering
     \includegraphics [width=\linewidth] {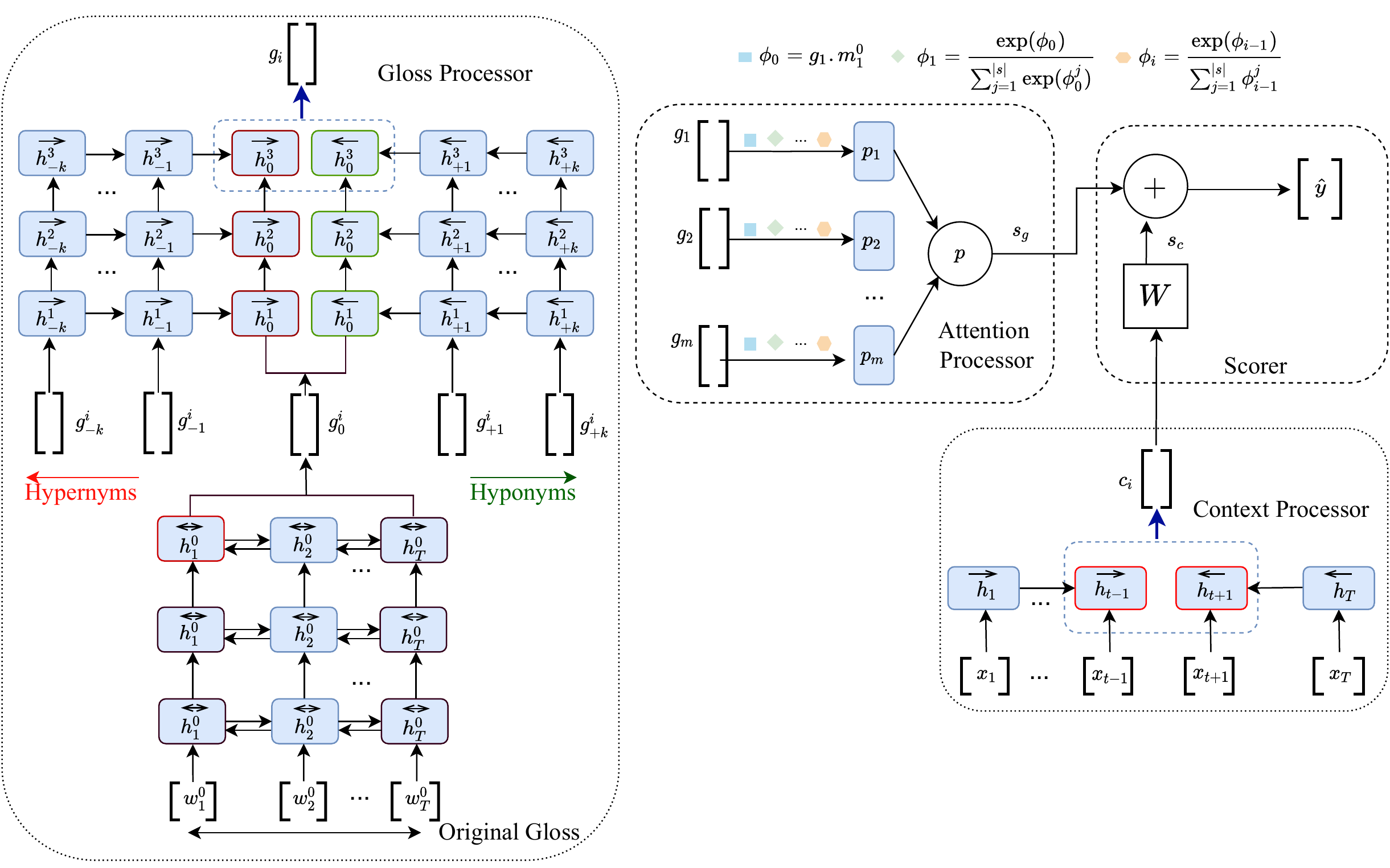}
    \caption{Overall architecture of \name~with SBiLSTM, Attention calculator, and Scorer components. }
    \label{main:fig}
\end{figure*}

In this section, we describe the proposed model Sense Picking for Word Sense Disambiguation (\name). 
The overall architecture of \name~consists of Stacked bi-directional LSTM (SBiLSTM), attention, and scoring components. 
Figure~\ref{main:fig} represents the overall architecture of \name~and corresponding workflow of individual components. 

\subsection{Stacked Bi-directional LSTM (SBiLSTM) Component}
 
Stacked bi-directional LSTM is implemented both for context and gloss processing as both of their data structures are sequential. Subsequently, in the figure, we depict two individual SBiLSTM architectures for processing context and gloss.

\subsubsection{Context Processor}

 \emph{Context processor} transforms the context of the target word into a distributed vector representation by considering a sequence of neighboring words.
The input to the context consists of the surrounding words $x_1, x_t,\dots, x_{t-1},x_{t+1},\dots, x_T$ of the target word, $x_t$ where $T$ denotes the total number of words present in the input sentence, $x$.
We use pre-trained word embedding matrix $M \in \mathbb{R}^{D\times V}$ to represent the surrounding words of $x_t$ in a $V$-dimensional vector as a one-hot encoding with $|D|$ unique words.
The forward LSTM takes the sequence of preceding words of $x_t$ represented as $x_1,x_2,\dots,x_{t-1}$ and computes a sequence of forward hidden states as $\overrightarrow{h}_1,\overrightarrow{h}_2,\dots,\overrightarrow{h}_{t-1}$.
Similarly, the backward LSTM takes the post sequence $x_T, \dots, x_{t+1}$ and computes a sequence of backward hidden states $\overleftarrow{h}_T,\dots,\overleftarrow{h}_{t+1}$.
Next, the hidden states are stacked before the final concatenation to get context, $c$ which is defined in Equation~\ref{eq:1},
\begin{equation}
\label{eq:1}
c=[\overrightarrow{h}_{t_1}:\overleftarrow{h}_{t+1}]
\end{equation}
 where : denotes concatenation operator.

\subsubsection{Gloss Processor}

We utilize the \emph{gloss processor} to transform each gloss into a vector.

First, gloss processor enriches the original gloss information, $g_o$ of a target word, $x_t$
by including its hypernyms $g_{\mu}$ and hyponyms $g_{\nu}$. Note that both hypernyms and hyponyms are obtained from WordNet.

Gloss set, $G$ holds the original gloss and the expansion (\ref{eq:g}).
 \begin{equation}
 \label{eq:g}
\centering
G =  g_o+ g_{\mu} + g_{\nu}
\end{equation}

In addition, we use tree depth $k$ to set limits in obtaining hypernyms and hyponyms.
$g_{\mu}$ and $g_{\nu}$ within $k$ depth are denoted as $[g-1, g-2, \dots, g-m]$ and $[g+1, g+2, \dots , g+n]$ respectively, where $m$ and $n$ denote number of hypernyms and hyponyms for the target word, $x_t$ within depth $k$. 
Thus, we determine $G$ for each $x_t$, denoted by $G_{x_t}$. 
Hence, we require to the process each gloss, $g^i_{x_t}$ of $G_{x_t}$ where $i \in [-m,-m+1, \dots , -1, 0, 1, \dots, n-1, n]$. 

Next, we use stacked bi-LSTM to encode each word $w_i$ of $g$ that is similar to processing context.
In addition, we use both $g_\mu$ and $g_\nu$ that are represented as concatenation of the respective word-embeddings from $M$.
A stacked forward LSTM is utilized to encode $g_{\mu}$ of $j$-th sense of $x_t$ that is $(\vec{g}^j_{-m}, \vec{g}^j_{-m+1}, \dots, \vec{g}^j_{1}, \vec{g}^j_{0})$  as a sequence of forward hidden states $(\vec{h}^j_{-m}, \vec{h}^j_{-m+1}, \dots, \vec{h}^j_{1}, \vec{h}^j_{0})$. 
Similarly, a stacked backward LSTM is leveraged to encode $g_{\nu}$ of $j$-th sense  
of $x_t$ that is $( {\cev{g}}^j_{-n+1}, \cev{g}^j_{-n+1}, \dots, \cev{g}^j_{1}, \cev{g}^j_{0}  )$  as a sequence of backward hidden states $(\cev{h}^j_{-n}, \cev{h}^j_{-n+1}, \dots, \cev{h}^j_{1}, \cev{h}^j_{0}  )$. 
For $j$-th sense of $x_t$, the resultant gloss vector, $g_j$ is the concatenation of the final state of the forward $\vec{h}^j_{0}$ and backward portions $\cev{h}^j_{0}$ and the vector is computed as
\begin{equation}
g_j = [\vec{h}^j_{0} : \cev{h}^j_{0}]   
\end{equation}

\subsection{Attention Processor}

The attention processor includes both an attention computation and memory update.
The component focuses on building the relationship between context and glosses by repeated attention generation.

First the attention $\phi$ for gloss $g$ is generated repeatedly at every iteration $i$ as follows

\begin{equation}
    \phi_i = f(g, m^{i-1}, c)
\end{equation}

\noindent where $m^{i-1}$ denotes the memory vector in the $(i-1)^{th}$ iteration and $c$ is the context component.
The scoring unit $f$ computes the semantic relationship between context and gloss components.
At the first iteration, the attention demonstrates the similarity of context and each gloss.
In the following iterations, the attention reflects the similarity between the updated context and a gloss.
The final attention $\phi_i$ of gloss $g$ at iteration $i$ is computed using dot product between gloss and context as

\begin{equation}
\phi_i= \frac{\exp(\phi_{i-1})} { \sum_{j=1}^{|s|} \exp(\phi_{i-1}^j) }
\end{equation}

\noindent where $\phi_0 = g_1 \cdot m_1^0$ and $|s|$ denotes the initial attention and  the number of word senses, respectively. 

Once we compute the final attention, we store the context state in $p^i$ by weighting the sum of glosses as $p_i = \sum_{i=1}^{n} \phi^i  g_i$ where $n$ denotes the hidden layer size of LSTM in both context and gloss components. Next, we update the context memory at $i$-th iteration using concatenation and bias $b$ using a rectified linear unit as follows.
\begin{equation}
 m^i = ReLU ( W[m^{i-1}:p^i:c] + b )   
\end{equation}
\noindent where : denotes concatenation operator and $b$ is bias.

\subsection{Scorer}
The scorer component computes the scores for the possible relevant senses with respect to $x_t$ and yields a probability distribution over all the senses. 
The overall score for each $g_j$ of $x_t$ is calculated by the latest attention $\phi_k$, where $k$ is the total number of passes in the attention component. 
The gloss score $s_g$ is computed by Equation~\ref{glos_att}.
\begin{equation}
\label{glos_att}
s_g=e_j^k 
\end{equation}
Where, $e^k$ denotes gloss score from $\phi_k$. 
On the other hand, we utilize a fully-connected layer to calculate the context score $s_c$ which is defined as $s_c=W_{x_t}c+b_{x_t}$ where $W_{x_t} \in \mathbb{R}^{s_n\times h_n}$, $b_{x_t} \in \mathbb{R}^{s_n}$, and $s_n$, $h_n$ denote total number of senses and hidden units in  forward and backward LSTM portions, respectively.
In addition, a learning parameter $\lambda$ is learned during the training phase. 
Using $\lambda$, we define Equation~\ref{eq:final} to estimate the sense probability distribution $\hat y$ over all of the senses with respect to $x_t$.
\begin{equation}
 \label{eq:final}
\hat{y}= S (\lambda_{x_t}\cdot s_c + \beta_{x_t} \cdot s_g)
\end{equation}
Where, $\beta_{x_t}=(1-\lambda_{x_t})$, both $\lambda_{x_t}$, $\beta_{x_t}$  $\in  [0,1]$, and $S(.)$ refers to softmax activation function.
For a given $x_t$, we aim to minimize the standard cross entropy loss between the actual sense (true label) y, and the predicted sense $\hat{y}$.

\section{Experiments and Evaluation}
The goal in this task is to disambiguate all the
content words in a given text corpus. To train
models for disambiguating a large set of content
words, a high-coverage sense-annotated corpus is
required. In this section, we describe the utilized benchmark datasets and compared the performance of \name~to other state-of-the-art methods. 

\subsection{Datasets}
\label{sec:dataset}
To build an effective WSD model, we require to utilize a high-coverage sense-annotated dataset. 
Hence, we utilize~\emph{SemCor 3.0} to train \name~since the dataset has been widely used in prior WSD modeling research~\cite{raganato2017neural,iacobacci2016embeddings}. The dataset is the largest manually sense-tagged corpus which contains $352$ number of documents within total $226,036$ senses. 

To evaluate \name, we utilize five benchmark \texttt{All-words} fine-grained datasets namely,~\textit{Senseval:02 (SE2), Senseval:03 task-1(SE3), SemEval:07 task-17 (SE7), SemEval:13 task-12 (SE13), and SemEval:15 task-13 (SE15)}. These datasets contain $7253$
sense tagged instances that are nouns, verbs, adjectives, and adverbs. We
utilize \textit{WordNet3.0} to obtain word sense glosses for input.

\subsection{Experimental Settings}
To obtain vector represenation of words, we use pre-trained word embeddings glove model containing $42$B words with $300$ dimensions and keep them fixed during the training process. We
assign gloss expansion depth $k=2$.  We employ $512$ hidden units in both
the context and gloss component for \name~and utilize Adam optimizer in the training process with $0.001$
initial learning rate, $\lambda$. To avoid the impact of overfitting,
we set dropout regularization and consider drop rate
as $0.5$. We train the model with $100$ epochs. Our implemented model is available in github~\footnote{\url{https://github.com/zobaed11/Spick.git}}.

\subsection{Compared Models}
\label{sec:ca}
We compare \name~with several state-of-the-art WSD models from knowledge-based, supervised, and neural network-based WSD domains.

\subsubsection{Knowledge-based}

\begin{itemize}
\item Enhanced Lesk~\cite{basile2014enhanced}: The model is an improved version of standard Lesk~\cite{lesk1986automatic}. The work considers overlapping of gloss and context to capture the importance of gloss for the disambiguation task. In addition, the authors expand the gloss set based on possible semantic relationship that is inferred from word embedding.  
    
    \item WN $1^{st}$ sense~\cite{raganato2017neural}: The model chooses the candidate sense that is appeared as first among the set of related senses for a particular word in \textit{WordNet 3.0}.   
    \item Babelfly~\cite{moro2015semeval}: This work is an unified graph-based models for the disambiguation task that leverages semantic network from \textit{BabelNet}.
    
    \item Baseline: Similar to prior research~\cite{raganato2017neural}, we employ a baseline model that chooses the most frequently annotated sense in the training dataset. 
    
\end{itemize}

\begin{table*}[!htbp]
\caption{F-1 score ($\%$) for English \texttt{All-words} datasets. The first five rows list knowledge-based models. Supervised and neural network-based models including \name, are listed in sixth-tenth rows. Bold-font indicates the top performer.}
\vspace{3 pt}
\centering
\begin{tabular}{|c|c|c|c|c|c|c|}
\hline
\textbf{Model Type} &
  \textbf{Related Work} &
  \textbf{SE2} &
  \textbf{SE3} &
  \textbf{SE7} &
  \textbf{SE13} &
  \multicolumn{1}{l|}{\textbf{SE15}} \\ \hline
\multirow{4}{*}{Knowledge-based} &
  Enhanced Lesk~\cite{basile2014enhanced} &
  63.0 &
  63.7 &
  56.7 &
  66.2 &
  64.6 \\ \cline{2-7} 
 &
  WN $1^{st}$ sense~\cite{raganato2017neural} &
  66.8 &
  66.2 &
  55.2 &
  63.0 &
  67.8 \\ \cline{2-7} 
 &
  Babelfly~\cite{moro2015semeval} &
  67.0 &
  63.5 &
  57.6 &
  66.2 &
  70.3 \\ \cline{2-7} 
 &
  Baseline &
  60.1 &
  62.3 &
  51.4 &
  63.8 &
  65.7 \\ \hline
\multirow{2}{*}{Supervised} &
  IMS~\cite{Zhong10h.t.:it} &
  70.9 &
  69.3 &
  58.5 &
  65.3 &
  69.5 \\ \cline{2-7} 
 &
  IMS$_{emb}$~\cite{iacobacci2016embeddings} &
  \textbf{72.2} &
  70.4 &
  61.2 &
  65.9 &
  71.5 \\ \hline
\multirow{7}{*}{Neural network-based} &
  WSD$_{BiLSTM}$~\cite{kaageback2016word} &
  71.1 &
  68.4 &
  61.5 &
  64.8 &
  68.3 \\ \cline{2-7} 
 &
  Context2Vec~\cite{melamud2016context2vec} &
  71.8 &
  69.1 &
  61.3 &
  65.6 &
  \textbf{71.9} \\ \cline{2-7} 
 &
  BiLSTM~\cite{raganato2017neural} &
  71.4 &
  68.8 &
  61.8 &
  65.4 &
  69.2 \\ \cline{2-7} 
 &
  BiLSTM + att. + LEX + POS~\cite{raganato2017neural} &
  72.0 &
  69.1 &
  62.8 &
  66.9 &
  71.5 \\ \cline{2-7} 
 &
Seq2Seq~\cite{raganato2017neural} &
  \multicolumn{1}{l|}{68.5} &
  \multicolumn{1}{l|}{67.9} &
  \multicolumn{1}{l|}{60.9} &
  {65.3} &
  {67.0} \\ \cline{2-7} 
 &
  GAS~\cite{luo2018incorporating} &
  72.0 &
  70.0 &
  63.2 &
 66.7 &
 71.6 \\ \cline{2-7} 
 &
  \emph{\name} &
  70.8 &
  \textbf{70.6} &
  \textbf{63.6} &
  \textbf{71.3} &
  70.9 \\ \hline
\end{tabular}
\label{table:all}
\end{table*}

\subsubsection{Supervised} 

\begin{itemize}
    \item \emph{It Makes Sense (IMS)}~\cite{Zhong10h.t.:it}: The model chooses Support Vector Machine (SVM) as the classifier and utilizes features (\ie POS tags, local co-occurrences) of the target word within a specific window. IMS trains a dedicated classifier for each word of the input sentence individually. 
    
    \item \emph{IMS$_{emb}$}~\cite{iacobacci2016embeddings}: The model uses IMS~\cite{Zhong10h.t.:it} as the framework but instead of utilizing local features (\ie POS tagging, local co-occurrences), Iacobacci \etal incorporate word embeddings as the features.  
\end{itemize}
    
    \subsubsection{Neural network-based}
    
    \begin{itemize}
        \item \emph{WSD$_{BiLSTM}$}~\cite{kaageback2016word}: The model leverages a bi-directional LSTM that is trained on an unlabeled dataset and shares model parameters among all words of input. However, the work does not consider the required gloss knowledge to build their model.

   \item \emph{BiLSTM + att. + LEX + POS}~\cite{raganato2017neural}: A bi-directional LSTM based model aims at modeling joint disambiguation of the target text as a sequence labeling problem.
    
    \item \emph{Context2Vec}~\cite{melamud2016context2vec}: Another bi-directional LSTM based work proposed by Melamud \etal that learns a context vector for each sense annotation in the training data.
    
    \item \emph{Sequence2sequence (Seq2Seq)}~\cite{raganato2017neural}: The model is based on encoder-decoder network that is widely adopted in NLP domain to perform various language modeling tasks in recent years. Raganato \etal propose the model with \emph{BiLSTM + att. + LEX + POS} to observe the performance of encoder-decoder architecture in sense disambiguation task.  
    
    \item \emph{Gloss augmented sens (GAS)}~\cite{luo2018incorporating}: The work also incorporates gloss information to train bi-directional LSTM neural model. They highlight the importance of incorporating expanded set of gloss in WSD task based on performance evaluation of their model. 
\end{itemize}

\subsection{Results and Discussion}
In this section, we depict the performance of \name~in the English \texttt{All-words} WSD task with respect to the models that are mentioned in Section~\ref{sec:ca}. 
Table~\ref{table:all} depicts the F1-score results on the five benchmark test datasets that is described in Section~\ref{sec:dataset}.

From the table, we observe that the supervised models outperform all knowledge-based models. Knowledge-based Babelfly achieves the overall best F-1 score among all knowledge-based models. In addition, we investigate that the inclusion of word embedding with the supervised approach improves performance. As a result, we observe that IMS$_{emb}$ outperforms the original IMS.
The underlying reason for the outperforming performance of IMS$_{emb}$ is the inclusion of word embedding that can reduce the negative effect of lower quality training data. 
Note that the supervised models train a dedicated classifier for each word of the input sentence individually that create a positive influence in sense prediction. Subsequently, it is a challenge for neural network-based models to beat supervised ones, and from the table, we observe that WSD$_{BiLSTM}$ and Seq2Seq models cannot beat supervised ones.

All neural network-based models clearly outperform all of the knowledge-based models for most of the datasets.     
In spite of having encoder-decoder architecture that is robust and powerful in language modeling, Seq2Seq model cannot achieve better result compared to other neural network-based models. We conclude that the variants of BiLSTM models perform better than encoder-decoder~\cite{devlin2018bert,laskar2020utilizing} network while performing the WSD task. The performance of the compared neural network-based models highlights that neural network is a good fit for disambiguation task as most of the compared neural network-based models perform closely and outperform other types of models.

Proposed \name~outperforms all supervised models as it beats both IMS and IMS$_{emb}$ on most of the datasets except SE2 and SE15.   
In comparison to the neural network-based models, \name~also clearly outperforms WSD$_{BiLSTM}$ and BiLSTM + att. + LEX + POS models on most of the datasets. Although Context2Vec and GAS models are the close competitors, \name~achieves the highest F-1 score in SE3, SE7, and SE13 datasets. 
From the table, we evaluate that \name~is the state-of-the-art performer considering all five test datasets and overall, \name~achieves a relative improvement of $3.5\%$ in F1 score. Incorporating glosses into the stacked BiLSTM-based model improves disambiguation performance.

\section{Conclusions}
In this research, we present \name, a unified neural model for the WSD task. We integrate the glosses knowledge of the ambiguous
word into a stacked bi-directional LSTM model.  
In this way, we not only utilize the sense-annotated context
data but also leverage the background knowledge (gloss) to obtain the appropriate word sense. Based on the experimental results on the English \texttt{All-words} WSD test datasets, we show that \name~beats other state-of-the-art performers in most of the datasets. We investigate that leveraging gloss information along with hypernyms and hyponyms and stacked BiLSTM architecture are the potential reasons for the efficacy of \name.

In future work, we plan to provide more empirical experiments for language independence by evaluating on several languages datasets.
We plan to extend this work by exploiting generative adversarial network (GAN)~\cite{goodfellow2014generative} to enrich the gloss set. This will help to reduce WordNet dependency as WordNet cannot provide gloss of every words, specifically the words that are domain-specific (\ie technical terms).

\bibliography{sample}

\begin{thebibliography}{10}

\bibitem{navigli2009word}
R.~Navigli, ``Word sense disambiguation: A survey,'' {\em Journal of ACM
  computing surveys (CSUR)}, vol.~41, no.~2, p.~10, 2009.

\bibitem{pal2019novel}
A.~R. Pal, D.~Saha, N.~S. Dash, S.~K. Naskar, and A.~Pal, ``A novel approach to
  word sense disambiguation in bengali language using supervised methodology,''
  {\em Journal of Sadhana}, vol.~44, no.~8, p.~181, 2019.

\bibitem{kaageback2016word}
M.~K{\aa}geb{\"a}ck and H.~Salomonsson, ``Word sense disambiguation using a
  bidirectional lstm,'' in {\em Proceedings of the 5th Workshop on Cognitive
  Aspects of the Lexicon (CogALex-V)}, pp.~51--56, 2016.

\bibitem{chaplot2018knowledge}
D.~S. Chaplot and R.~Salakhutdinov, ``Knowledge-based word sense disambiguation
  using topic models,'' in {\em Proceedings of the 32nd AAAI Conference on
  Artificial Intelligence}, 2018.

\bibitem{lesk1986automatic}
M.~Lesk, ``Automatic sense disambiguation using machine readable dictionaries:
  how to tell a pine cone from an ice cream cone,'' in {\em Proceedings of the
  5th annual international conference on Systems documentation}, pp.~24--26,
  Citeseer, 1986.

\bibitem{tang2010semi}
X.~Tang, X.~Chen, W.~Qu, and S.~Yu, ``Semi-supervised wsd in selectional
  preferences with semantic redundancy,'' in {\em Proceedings of the 23rd
  International Conference on Computational Linguistics: Posters},
  pp.~1238--1246, Association for Computational Linguistics, 2010.

\bibitem{karim2020efficient}
A.~Karim, S.~Azam, B.~Shanmugam, and K.~Kannoorpatti, ``Efficient clustering of
  emails into spam and ham: The foundational study of a comprehensive
  unsupervised framework,'' {\em Journal of IEEE Access}, vol.~8,
  pp.~154759--154788, 2020.

\bibitem{chen2013applying}
Y.~Chen, H.~Cao, Q.~Mei, K.~Zheng, and H.~Xu, ``Applying active learning to
  supervised word sense disambiguation in medline,'' {\em Journal of the
  American Medical Informatics Association}, vol.~20, no.~5, pp.~1001--1006,
  2013.

\bibitem{miller1995wordnet}
G.~A. Miller, ``Wordnet: a lexical database for english,'' {\em Journal of
  Communications of the ACM}, vol.~38, no.~11, pp.~39--41, 1995.

\bibitem{navigli2012babelnet}
R.~Navigli and S.~P. Ponzetto, ``Babelnet: The automatic construction,
  evaluation and application of a wide-coverage multilingual semantic
  network,'' {\em Journal of Artificial Intelligence}, vol.~193, pp.~217--250,
  2012.

\bibitem{basile2014enhanced}
P.~Basile, A.~Caputo, and G.~Semeraro, ``An enhanced lesk word sense
  disambiguation algorithm through a distributional semantic model,'' in {\em
  Proceedings of COLING 2014, the 25th International Conference on
  Computational Linguistics: Technical Papers}, pp.~1591--1600, 2014.

\bibitem{gers2000learning}
F.~A. Gers, J.~Schmidhuber, and F.~Cummins, ``Learning to forget: Continual
  prediction with lstm,'' {\em Journal of Neural Computation}, vol.~12, no.~10,
  pp.~2451--2471, 2000.

\bibitem{camacho2016large}
J.~Camacho-Collados, C.~D. Bovi, A.~Raganato, and R.~Navigli, ``A large-scale
  multilingual disambiguation of glosses,'' in {\em Proceedings of the Tenth
  International Conference on Language Resources and Evaluation (LREC'16)},
  pp.~1701--1708, 2016.

\bibitem{guo2010combining}
W.~Guo and M.~Diab, ``Combining orthogonal monolingual and multilingual sources
  of evidence for all words wsd,'' in {\em Proceedings of the 48th Annual
  Meeting of the Association for Computational Linguistics}, pp.~1542--1551,
  Association for Computational Linguistics, 2010.

\bibitem{moro2015semeval}
A.~Moro and R.~Navigli, ``Semeval-2015 task 13: Multilingual all-words sense
  disambiguation and entity linking,'' in {\em Proceedings of the 9th
  international workshop on semantic evaluation (SemEval 2015)}, pp.~288--297,
  2015.

\bibitem{brin1998anatomy}
S.~Brin and L.~Page, ``The anatomy of a large-scale hypertextual web search
  engine,'' {\em Journal of Computer networks and ISDN systems}, vol.~30,
  no.~1-7, pp.~107--117, 1998.

\bibitem{correa2018word}
E.~A. Correa~Jr, A.~A. Lopes, and D.~R. Amancio, ``Word sense disambiguation: A
  complex network approach,'' {\em Journal of Information Sciences}, vol.~442,
  pp.~103--113, 2018.

\bibitem{mikolov2013efficient}
T.~Mikolov, K.~Chen, G.~Corrado, and J.~Dean, ``Efficient estimation of word
  representations in vector space,'' {\em arXiv preprint arXiv:1301.3781},
  2013.

\bibitem{pennington2014glove}
J.~Pennington, R.~Socher, and C.~D. Manning, ``Glove: Global vectors for word
  representation,'' in {\em Proceedings of the 2014 conference on empirical
  methods in natural language processing (EMNLP)}, pp.~1532--1543, 2014.

\bibitem{devlin2018bert}
J.~Devlin, M.-W. Chang, K.~Lee, and K.~Toutanova, ``Bert: Pre-training of deep
  bidirectional transformers for language understanding,'' {\em arXiv preprint
  arXiv:1810.04805}, 2018.

\bibitem{luo2018leveraging}
F.~Luo, T.~Liu, Z.~He, Q.~Xia, Z.~Sui, and B.~Chang, ``Leveraging gloss
  knowledge in neural word sense disambiguation by hierarchical co-attention,''
  in {\em Proceedings of the 2018 Conference on Empirical Methods in Natural
  Language Processing}, pp.~1402--1411, 2018.

\bibitem{raganato2017neural}
A.~Raganato, C.~D. Bovi, and R.~Navigli, ``Neural sequence learning models for
  word sense disambiguation,'' in {\em Proceedings of the 2017 Conference on
  Empirical Methods in Natural Language Processing}, pp.~1156--1167, 2017.

\bibitem{luo2018incorporating}
F.~Luo, T.~Liu, Q.~Xia, B.~Chang, and Z.~Sui, ``Incorporating glosses into
  neural word sense disambiguation,'' in {\em Proceedings of the 56th Annual
  Meeting of the Association for Computational Linguistics (Volume 1: Long
  Papers)}, pp.~2473--2482, 2018.

\bibitem{yang2019leveraging}
B.~Yang and T.~Mitchell, ``Leveraging knowledge bases in lstms for improving
  machine reading,'' in {\em Proceedings of the 55th Annual Meeting of the
  Association for Computational Linguistics (Volume 1: Long Papers)},
  pp.~1436--1446, 2017.

\bibitem{wang2016dimensional}
J.~Wang, L.-C. Yu, K.~R. Lai, and X.~Zhang, ``Dimensional sentiment analysis
  using a regional cnn-lstm model,'' in {\em Proceedings of the 54th Annual
  Meeting of the Association for Computational Linguistics (Volume 2: Short
  Papers)}, pp.~225--230, 2016.

\bibitem{baziotis2017datastories}
C.~Baziotis, N.~Pelekis, and C.~Doulkeridis, ``Datastories at semeval-2017 task
  4: Deep lstm with attention for message-level and topic-based sentiment
  analysis,'' in {\em Proceedings of the 11th International Workshop on
  Semantic Evaluation (SemEval-2017)}, pp.~747--754, 2017.

\bibitem{song2019abstractive}
S.~Song, H.~Huang, and T.~Ruan, ``Abstractive text summarization using lstm-cnn
  based deep learning,'' {\em Journal of Multimedia Tools and Applications},
  vol.~78, no.~1, pp.~857--875, 2019.

\bibitem{liu2018generative}
L.~Liu, Y.~Lu, M.~Yang, Q.~Qu, J.~Zhu, and H.~Li, ``Generative adversarial
  network for abstractive text summarization,'' in {\em Thirty-second AAAI
  conference on artificial intelligence}, 2018.

\bibitem{melamud2016context2vec}
O.~Melamud, J.~Goldberger, and I.~Dagan, ``context2vec: Learning generic
  context embedding with bidirectional lstm,'' in {\em Proceedings of the 20th
  SIGNLL conference on computational natural language learning}, pp.~51--61,
  2016.

\bibitem{yuan2016semi}
D.~Yuan, J.~Richardson, R.~Doherty, C.~Evans, and E.~Altendorf,
  ``Semi-supervised word sense disambiguation with neural models,'' in {\em
  Proceedings of COLING 2016, the 26th International Conference on
  Computational Linguistics: Technical Papers}, pp.~1374--1385, 2016.

\bibitem{rumelhart1986learning}
D.~E. Rumelhart, G.~E. Hinton, and R.~J. Williams, ``Learning representations
  by back-propagating errors,'' {\em Journal of nature}, vol.~323, no.~6088,
  pp.~533--536, 1986.

\bibitem{pascanu2014construct}
R.~Pascanu, C.~Gulcehre, K.~Cho, and Y.~Bengio, ``How to construct deep
  recurrent neural networks,'' in {\em Proceedings of the Second International
  Conference on Learning Representations (ICLR 2014)}, 2014.

\bibitem{iacobacci2016embeddings}
I.~Iacobacci, M.~T. Pilehvar, and R.~Navigli, ``Embeddings for word sense
  disambiguation: An evaluation study,'' in {\em Proceedings of the 54th Annual
  Meeting of the Association for Computational Linguistics (Volume 1: Long
  Papers)}, pp.~897--907, 2016.

\bibitem{Zhong10h.t.:it}
Z.~Zhong and H.~T. Ng, ``H.t.: It makes sense: A wide-coverage word sense
  disambiguation system for free text,'' in {\em Proceedings of the 48th Annual
  Meeting of the Association for Computational Linguistics (ACL}, pp.~78--83,
  2010.

\bibitem{laskar2020utilizing}
M.~T.~R. Laskar, E.~Hoque, and J.~X. Huang, ``Utilizing bidirectional encoder
  representations from transformers for answer selection,'' {\em arXiv preprint
  arXiv:2011.07208}, 2020.

\bibitem{goodfellow2014generative}
I.~Goodfellow, J.~Pouget-Abadie, M.~Mirza, B.~Xu, D.~Warde-Farley, S.~Ozair,
  A.~Courville, and Y.~Bengio, ``Generative adversarial nets,'' in {\em
  Advances in neural information processing systems}, pp.~2672--2680, 2014.

\end{thebibliography}
\bibliographystyle{ieeetr}

\end{document}